\documentclass[sigconf]{acmart}

\usepackage{booktabs} 
\usepackage{color}
\usepackage{caption}
\usepackage{subcaption}

\copyrightyear{2017} 
\acmYear{2017}

\begin{document}

\acmPrice{15.00}
\acmDOI{10.1145/3133944.3133945}
\acmConference{AVEC'17}{October 23, 2017} {Mountain View, CA, USA.}

\acmISBN{978-1-4503-5502-5/17/10}

\title{ Topic Modeling Based Multi-modal Depression Detection }

\author{Yuan Gong}
\affiliation{\institution{University of Notre Dame}}
\email{ygong1@nd.edu}

\author{Christian Poellabauer}
\affiliation{\institution{University of Notre Dame}}
\email{cpoellab@cse.nd.edu}

\begin{abstract}

Major depressive disorder is a common mental disorder that affects almost 7\% of the adult U.S. population. The 2017 Audio/Visual Emotion Challenge (AVEC) asks participants to build a model to predict depression levels based on the audio, video, and text of an interview ranging between 7-33 minutes. Since averaging features over the entire interview will lose most temporal information, how to discover, capture, and preserve useful temporal details for such a long interview are significant challenges. Therefore, we propose a novel topic modeling based approach to perform context-aware analysis of the recording. Our experiments show that the proposed approach outperforms context-unaware methods and the challenge baselines for all metrics.

\end{abstract}


\maketitle

\section{Introduction}

\subsection{Background}

Major depressive disorder (MDD), also usually called depression, is one of the most common mood disorders, which is characterized by a persistent low mood. The study in~\cite{fava2000major} showed that men have a risk of 10-20\% and women have a risk of 5-12\% to develop MDD in their lifetime. Early and accurate detection of MDD will ensure that appropriate treatment and intervention options can be considered. Therefore, there is a strong need for a simple method to detect depression. In the 2017 Audio/Visual Emotion Challenge (AVEC)~\cite{avec2017baseline}, the depression sub-challenge task requires participants to predict the depression level (i.e., the PHQ-8 score~\cite{kroenke2009phq}) using audio, video, and text analysis. The database used in this challenge is the distress analysis interview corpus (DAIC-WOZ) ~\cite{gratch2014distress},~\cite{devault2014simsensei}, which includes data from 189 subjects. For each subject, the database includes the audio/video features as well as the transcript of an interview ranging between 7-33 minutes, which is conducted by an animated virtual interviewer called Ellie, controlled by a human interviewer in another room.

\begin{figure}[htp]
  \centering
  \includegraphics[width=8.4cm]{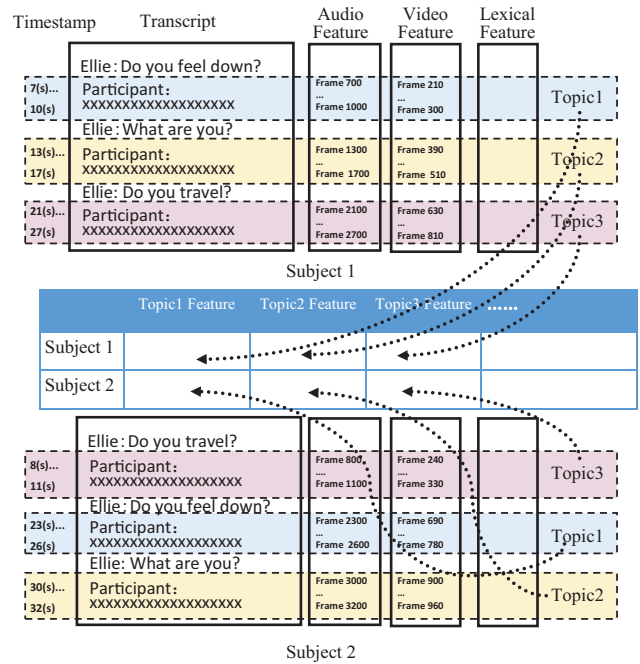}
  \caption{Illustration of the proposed topic modeling based multi-modal feature vector building scheme.}
  \label{fig:ilus}
\end{figure}

\subsection{Challenges and Contribution}
A big difference between the depression detection task and a traditional emotion detection task is the decision unit. Since human emotion can change rather quickly, traditional emotion detection typically requires second-level prediction. Therefore, popular emotion recognition databases usually provide labels for short-term recordings, e.g., the IEMOCAP database~\cite{busso2008iemocap} provides labels for each utterance, while the SEWA database provides labels for each segment of 100ms. In contrast, depression is expressed through a persistently low mood, which is very different from short-term sadness. The study in~\cite{spijker2002duration} shows that the median duration of depression is three months and consequently, prediction of the depression level of an individual should be based on much longer observation periods.
This difference between depression and emotion detection leads to two main challenges:
\begin{enumerate}
\item{\bf Large decision unit.} In the DAIZ-WOZ database, each data sample is the audio and video recording of an interview of a specific subject, where the interview ranges from 7 to 33 minutes. Here, only one decision needs to be made for the entire interview. The length of the decision unit is much longer than for typical emotion recognition tasks, e.g., the 2017 AVEC emotion sub-challenge requires making decisions for each 100ms segment. While a large data volume is typically beneficial for the accuracy, processing large amounts of data can be challenging. When analyzing very long audio/video data, applying statistical functions (e.g., max, min, mean, quartiles) to short-term features over the entire interview will lead to loss of detailed temporal information such as short-term sighs in despair, laughing, or anger. However, these short-term details within the interview can be useful when determining the depression level of the subject, especially when analyzed together with contextual information (e.g., sighing in despair when being asked about sleep quality, laughing when talking about a journey, and anger when remembering unhappy experiences). Therefore, it is important to map the whole interview to a feature vector such that short-term details and context are maintained.

\item {\bf Limited number of samples.} Since each subject has one sample (the entire audio/video recording), the number of samples is significantly lower than in the case when each recording consists of many small samples (e.g., each utterance being a sample). In the 2017 AVEC depression sub-challenge, the number of samples in the training set is only 107. In addition, the database is unevenly distributed, i.e., the number of depression samples in the training set is 30. With such a small sample size, the number of features should also be small to avoid the problems of dimensionality and overfitting. However, the dimensions of audio and video features are very large and therefore, generating and selecting an appropriate number of discriminative features is essential.

\end{enumerate}

In order to overcome these two challenges, we propose a topic modeling based multi-modal feature vector building scheme as shown in Figure~\ref{fig:ilus} to provide the basis for context-aware analysis. The interview is first segmented according to topics. Then, audio, video, and semantic features are generated for each topic segment separately and further placed into a separate slot of the topic in the feature vector. After the features for all topics have been placed, a two-step feature selection algorithm is executed to shrink the feature vector and only keep the most discriminative features. The proposed algorithm is inspired by the observation that all interviews contain not a fixed, but a limited range of topics. Further, we assume that each question by Ellie triggers a response on a new topic, which makes ``topic tracking'' feasible. We expect the following advantages from the proposed scheme:
\begin{enumerate}
\item{\bf Logically organize short-term details based on context.} When retaining the short-term details of the interview, we need to do it in a fashion that keeps the feature vector space relatively small and also makes it logical. Extracting details according to utterances is not context-organizable and will lead to a dimension explosion since each interview contains hundreds of utterances. For example, both subject 1 and subject 2 smile at utterance 10, but their smiling might convey different information since their 10th utterance is in different contexts. In contrast, the proposed scheme tracks the topic and place feature of utterances, no matter where it is in the interview, into the slot of the topic it belongs to in the feature vector. In addition, one topic can cover multiple utterances, which makes the feature dimension much smaller. 
\item {\bf More flexible and precise discovery of useful features.} In traditional feature building schemes, one feature can only be kept or discarded as a whole. However, it is common that one feature is only useful in some specific contexts and useless in others. Further, same features in different contexts might convey different information and should be regarded as separate features. For example, smiling in the context of discussing family can be more discriminative than smiling in the context of greeting someone, because the latter might only be due to etiquette. Therefore, we would like to only keep the feature when it is in a useful context. The proposed feature building scheme provides any combination of features and contexts such as smiling (family) and smiling (greeting). Thus, the feature selection algorithm can perform a more flexible and accurate filtering. In summary, the proposed scheme allows us to perform a finer analysis of the subject's reaction to a specific topic, such as a lower voice when discussing family, irritation when discussing an unhappy situation, and the expressions used when describing recent emotions. We believe that this finer-grained analysis can improve the performance of depression detection.
\end{enumerate}






\subsection{Related Work}

In the 2016 AVEC depression classification sub-challenge~\cite{valstar2016avec}, a few proposed techniques adopted text analysis for their model building. In~\cite{pampouchidou2016depression} and~\cite{nasir2016multimodal}, the text is analyzed on a subject level and audio/video features are separately extracted and then fused with semantic features, i.e., topic modeling is not used in these approaches. In~\cite{williamson2016detecting}, the authors conduct a question/answer extraction (which is similar to topic extraction) before text analysis. However, the question/answer extraction is only applied to text analysis. Audio and video analysis is still conducted separately. In~\cite{yang2016decision}, the authors also conduct topic extraction, but merely use the semantic features of very few topics (3 topics for women and 4 topics for men) to build a simple decision tree. This approach achieved the best performance in the 2016 AVEC, which demonstrates the effectiveness of simple model and key topic analysis. However, its performance on the test set is much worse than that on the development set. Its limited ability to generalize is probably due to the very small number of features. Further, audio and video features are not used in this work. 

On the other hand, topic modeling, which is a technique to discover topics from documents, has been widely adopted in applications such as text mining~\cite{hong2010empirical} and recommendation systems~\cite{wang2011collaborative}. It recently has also been used for depression and neuroticism assessment~\cite{resnik2013using}. In~\cite{resnik2013using}, the authors demonstrate that taking automatically derived topics into account improves prediction performance. However, audio and video analysis are again not involved in this work.

In summary, the work in~\cite{williamson2016detecting}, ~\cite{yang2016decision}, and~\cite{resnik2013using} use topic modeling, but only for text analysis. We further extend the application of topic modeling by using it for context-aware audio and video analysis. To the best of our knowledge, the proposed work is the first effort to combine topic modeling with multi-modal text, audio, and video analysis. 

\begin{table*}[]
\small 
\centering
\caption{The list of topics extracted from the DAIC-WOZ (potential key topics are marked with an asterisk).}
\label{tab:topicList}
\begin{tabular}{@{}p{0.2cm}p{2cm}p{5.5cm}p{0.3cm}p{2cm}p{5.5cm}@{}}
\toprule
Ind. & Topic Abbr.            & Sample Ellie Question                                                              & Ind. & Topic Abbr.       & Sample Ellie Question                                                       \\ \midrule
1     & more                           & can you tell me about that                                                         & 43    & best\_parent             & what's the best thing about being a parent                                  \\
2     & why                            & why                                                                                & 44    & are\_you\_okay           & are you okay with this                                                      \\
3     & last\_happy\_time              & tell me about the last time you felt really happy                                  & 45    & mad                      & what are some things that make you really mad                               \\
4     & origin                         & where are you from originally                                                      & 46    & they\_triggered          & are they triggered by something                                             \\
5     & argue                          & when was the last time you argued with someone and what was it about               & 47    & easy\_parent             & do you find it easy to be a parent                                          \\
6     & advice\_ago                    & what advice would you give to yourself ten or twenty years ago                     & 48    & happy\_did\_that         & are you happy you did that                                                  \\
7     & control\_temper                & how are you at controlling your temper                                             & 49    & therapist\_affect        & how has seeing a therapist affected you                                     \\
8     & things\_like\_la               & what are some things you really like about l\_a                                    & 50    & job                      & what are you                                                                \\
9     & proud                          & what are you most proud of in your life                                            & 51    & symptoms                 & what were your symptoms                                                     \\
10    & positive\_influence   & who's someone that's been a positive influence in  your life                       & 52    & ideal\_weekend           & tell me how you spend your ideal weekend                                    \\
11    & best\_friend\_describe         & how would your best friend describe you                                            & 53    & avoid                    & could you have done anything to avoid it                                    \\
12    & things\_dont\_like\_la         & what are some things you don't really like about l\_a                              & 54    & do\_annoyed              & what do you do when you are annoyed                                         \\
13    & major                          & what did you study at school                                                       & 55    & got\_in\_trouble         & has that gotten you in trouble                                              \\
14    & regret                         & is there anything you regret                                                       & 56    & your\_kid                & tell me about your kids                                                     \\
15    & dream\_job                     & what's your dream job                                                              & 57    & someone\_made\_bad       & tell me about a time when someone made you feel really badly about yourself \\
16    & enjoy\_travel                  & what do you enjoy about traveling                                                  & 58    & different\_parent        & what are some ways that you're different as a parent than your parents      \\
17    & how\_hard                      & how hard is that                                                                   & 59    & today\_kid               & what do you think of today's kids                                           \\
18    & do\_sleep\_not\_well           & what are you like when you don't sleep well                                        & 60    & down                     & do you feel down                                                            \\
19    & experiences         & what's one of your most memorable experiences                                      & 61    & how\_know\_them          & how do you know them                                                        \\
20    & hardest\_decision              & tell me about the hardest decision you've ever had to make                         & 62    & feel\_often              & do you feel that way often                                                  \\
21    & fun\_relax                     & what are some things you like to do for fun                                        & 63    & problem\_before          & did you think you had a problem before you found out                        \\
22    & handle\_differently & tell me about a situation that you wish you had handled differently                & 64    & living\_situation        & how do you like your living situation                                       \\
23    & what\_decide                   & what made you decide to do that                                                    & 65    & why\_stop                & why did you stop                                                            \\
24    & still\_work                    & are you still doing that                                                           & 66    & how\_do\_you\_do         & how are you doing today                                                     \\
25    & erase\_memory                  & tell me about an event or something that you wish you could erase from your memory & 67    & roommate                 & do you have roommates                                                       \\
26    & why\_move\_la                  & why did you move to l\_a                                                           & 68    & hard\_on\_yourself       & do you think that maybe you're being a little hard on yourself              \\
27    & change\_self                   & what are some things you wish you could change about yourself                      & 69    & like\_living\_with       & what's it like for you living with them                                     \\
28    & best\_quality                  & what would you say are some of your best qualities                                 & 70    & disturb\_thought         & do you have disturbing thoughts                                             \\
29    & often\_back                    & how often do you go back to your home town                                         & 71    & where\_live              & where do you live                                                           \\
30    & how\_long\_diagnose            & how long ago were you diagnosed                                                    & 72    & after\_millitary         & what did you do after the military                                          \\
31    & guilty                         & what's something you feel guilty about                                             & 73    & combat                   & did you ever see combat                                                     \\
32    & when\_move\_la                 & when did you move to l\_a                                                          & 74    & talk\_later              & why don't we talk about that later                                          \\
33    & easy\_used\_la                 & how easy was it for you to get used to living in l\_a                              & 75    & military\_change         & how did serving in the military change you                                  \\
34    & seek\_help                     & what got you to seek help                                                          & 76*    & change\_behavior         & have you noticed any changes in your behavior or thoughts lately            \\
35    & when\_last\_happy              & when was the last time you felt really happy                                       & 77*    & depression               & have you been diagnosed with depression                                     \\
36    & cope                           & how do you cope with them                                                          & 78*    & easy\_sleep              & how easy is it for you to get a good night sleep                            \\
37    & compare\_la                    & how does it compare to l\_a                                                        & 79*    & family\_close            & how close are you to your family                                            \\
38    & hard\_parent                   & what's the hardest thing about being a parent                                      & 80*    & feeling\_lately          & how have you been feeling lately                                            \\
39    & still\_therapy                 & do you still go to therapy now                                                     & 81*    & shy\_outgoing & do you consider yourself an introvert                                       \\
40    & travel\_a\_lot                 & do you travel a lot                                                                & 82*    & ptsd                     & have you ever been diagnosed with p\_t\_s\_d                                \\
41    & ever\_served\_military         & have you ever served in the military                                               & 83*    & therapy\_useful          & do you feel like therapy is useful                                          \\
42    & when\_last\_time               & when was the last time that happened                                               &       &                          &                                                                             \\ \bottomrule
\end{tabular}
\end{table*}

\section{Topic modelling based multi-modal depression detection}
\label{sec:proposed}

\subsection{Topic Modeling}
\label{sec:topicModel}

Topic modeling typically requires a sophisticated algorithm such as latent Dirichlet allocation (LDA)~\cite{blei2003latent} and network regularization~\cite{mei2008topic}. However, for the transcription of clinical interviews (such as provided by the DAIC-WOZ database), topic modeling can be done much simpler for multiple reasons. First, in the interview, only Ellie determines the topic by asking a question and the subject does not initiate a topic proactively. Second, the number of topics in clinical interviews is limited. And third, Ellie is an animated interviewer controlled by human command and, therefore, has a relatively fixed way to start a topic. We observed that when starting a specific topic, Ellie chooses one sentence from the library, which typically consists of only 1-3 fixed sentences per topic. 

Based on the above-mentioned observations, we perform simple topic modeling on the text of interviews. First, we build a preliminary sentence dictionary by traversing all of Ellie's speech and record all non-redundant sentences. Then, we perform manual cleaning of the preliminary dictionary, where sentences that do not start new topics (e.g., ``that's good'') are discarded.
After that, we perform clustering of the dictionary, where the sentences that start the same topic are grouped together. This is done in two steps. First, very similar sentences with up to 3 characters difference are clustered automatically. Second, further manual clustering and checks are performed. Then, we review each sentence cluster, link each cluster to the corresponding topic, and put it into the topic dictionary. Therefore, the topic dictionary is formatted as [topic name, corresponding Ellie sentences]. The complete list with 83 extracted topics is shown in Table~\ref{tab:topicList}. 

\begin{figure}[htp]
  \centering
  \includegraphics[width=8.4cm]{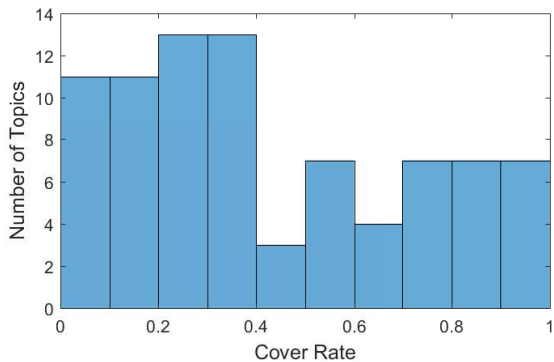}
  \caption{Histogram of topic cover rate.}
  \label{fig:sparsity}
\end{figure}

Note that only a few topics are discussed in most interviews, e.g., only 14 topics cover over 80\% of the interviews. In other words, topics are sparsely distributed in the interviews. The histogram of the topic cover rate is shown in Figure~\ref{fig:sparsity}.

\subsection{Features}
\label{sec:feature}

In this work, we use audio, video, and semantic features to build a multi-modal model. Audio and video features are provided by the 2017 AVEC organizers while semantic features are extracted by ourselves. All features are computed in a topic-wise fashion.

\subsubsection{Audio Features}

We use the audio features extracted by the COVAREP toolkit~\cite{degottex2014covarep} and formant features. The COVAREP toolkit generates a 74-dimensional feature vector that includes common features such as fundamental frequency and peak slope. Formant features contain the first 5 formants,
i.e., the vocal tract resonance frequencies. Both COVAREP and formant features are extracted every 10ms. For each topic, we further apply three statistic functions (mean, max, and min) to each feature over time to reduce the dimension. That is, for each topic, $(74+5)\times3=237$ audio features are used.

\subsubsection{Video Features}

We use the action units (AUs) features extracted by the OpenFace toolkit~\cite{baltruvsaitis2016openface}, which includes the information of 20 key AUs. For each topic, we further apply three statistic functions (mean, max, and min) over time to each feature to reduce the dimension. Thus, for each topic, $20\times3=60$ video features are used.

\subsubsection{Semantic Features}

For each of the 83 topics, we use the Linguistic Inquiry and Word Count (LIWC)~\cite{pennebaker2015development} software to count the frequency of word presence of the subject's speech of the topic in 93 categories such as anger, negative emotion, and positive emotion. That is, the LIWC software takes the speech of a subject and generates a 93-dimensional feature vector. 
Further, inspired by~\cite{yang2016decision}, which demonstrates that some key topics such as sleep quality (topic index: 78) and PTSD diagnose history (topic index: 82) have a high correlation with depression level, we further extract additional semantic features for 8 topics (topic index: 76-83, marked with an asterisk in Table~\ref{tab:topicList}) that we believe might be most discriminative. We use a dictionary based method to classify each topic into 2 or 3 categories according to the content. For example, for the topic easy\_sleep (topic index: 78), the speech of each subject is classified into three categories: easy (when phrases such as `no problem' are present), fair (when words such as `it depends' are present), and hard (when words such as `difficult' are present). The dictionary is built manually for each key topic.

\begin{figure}[htp]
  \centering
  \includegraphics[width=8.4cm]{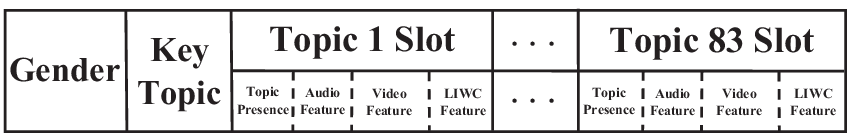}
  \caption{Illustration of the structure of feature vector.}
  \label{fig:featureVec}
\end{figure}

\begin{table}[]
\centering
\caption{Dimension of each feature category.}
\label{tab:dim}
\begin{tabular}{@{}p{3cm}c@{}}
\toprule
Feature Name       & Dimension \\ \midrule
Gender             & 1         \\
Topic Presence     & 83        \\
Key Topic & 8         \\
LIWC               & 7719      \\
Formant            & 1245      \\
COVAREP            & 18426     \\
AUs                & 4980      \\\midrule
Sum                & 32462     \\ \bottomrule
\end{tabular}
\end{table}

\subsection{Topic-wise Feature Mapping}
\label{sec:featureVector}

In order to conduct context-aware analysis, the feature vector needs to record the features of each topic separately. Therefore, in the feature vector, each topic has a separate slot.

We first find the topics discussed in each interview. For each interview, speech sentences of Ellie are traversed and when the sentence is found in the topic dictionary, the corresponding topic and the subject's speech, together with its timestamps are recorded. The subject's speech is used to generate semantic features while the timestamps are used to synchronize audio and video features. Then, all features are placed into separate slots of the corresponding topic in the feature vector. As described in Section~\ref{sec:feature}, each topic contains 237 audio features, 60 video features, and 93 LIWC features, and there are 83 topics in total, which leads to a $83*(237+60+93)=32,370$ dimensional feature vector. Further, we add the presence of each topic to the feature vector, because each interview only covers a few topics and the topic presence might be correlated to the subject's status. Finally, gender is also attached to the feature vector similar to the work in~\cite{yang2016decision} and~\cite{pampouchidou2016depression}, where the authors report that gender information can greatly improve the classification performance. Figure~\ref{fig:featureVec} illustrates the structure of the feature vector and Table~\ref{tab:dim} shows the dimension of each feature category in the feature vector. Due to the sparsity of topics, the feature vector is also sparse, i.e., the features of topics that are not discussed in an interview are missing. However, the slots for all topics are preserved in our approach, i.e., the slot of a topic that is not discussed in the interview is padded with -1. 

\subsection{Feature Selection}
\label{sec:featureSelection}

In Section~\ref{sec:featureVector}, a 32,462-dimensional feature vector is built, which maintains audio, video, and text information of each topic. However, only a small amount of features are actually useful and we expect the number of features to be small enough to avoid potential overfitting. Therefore, feature selection is an essential step of the proposed scheme.

We conduct feature selection in two steps. First, we conduct a quick model-independent feature selection on all features. The algorithm we use in this step is correlation-based feature subset selection (CFS)~\cite{hall1998correlation}, which evaluates the value of a subset of features by considering the individual predictive ability of each feature along with the degree of redundancy between them. After this step, a subset of features is selected. Then, we conduct a fine model-dependent feature selection to find the optimized feature number. In this step, we first rank the features according to their F-value to the corresponding label. Then we run the regression algorithm using a various number of high-rank features and finally select the best feature set.

This unique two-step feature selection algorithm is designed based on the following consideration. In our feature generation scheme, we observe that more features are correlated to each other than with the context-unaware feature generation scheme, because features that belong to the same topic are likely to have high correlations. Thus, if we only conduct feature selection according to the individual feature score, we might get a set of features with high scores, but that are also closely correlated to each other. In other words, many features are redundant and provide little extra information in this case. To avoid that, we first conduct a CFS to select a feature subset, where features have a high correlation with the label, but low correlation with each other. Since CFS is a model-independent approach, which cannot tell us the overfitting risk for our specific model and dataset, we further conduct a model-based selection on our dataset to find the appropriate number of features for our task. 

\subsection{Regression Model Building}

\subsubsection{Data Balancing}
It has been widely reported that imbalanced classes of data will greatly affect the performance of machine learning algorithms~\cite{liu2009exploratory}. Unfortunately, most healthcare related databases, including DAIZ-WOZ, are imbalanced. In the training set of the DAIC-WOZ database, only 30 subjects are depressed of a total of 107 subjects, which means that there are much more subjects with low PHQ-8 scores than those with high PHQ-8 scores. Therefore, we perform random-oversampling to make the number of samples for each PHQ-8 score is roughly the same by simply duplicating samples before running the machine learning algorithm. 

\subsubsection{Regressors}

In this work, we perform a grid search for the following regression models: random forest regression (number of trees: 1, 10, 20, 30, 40, 50, 100, and 200), stochastic gradient descent (SGD) regression, and support vector regression (SVR) (kernel: linear, polynomial, and radial basis function (RBF)).

\section{Experimental Setup}

\subsection{Test Strategy}
\label{sec:testStrategy}

In the 2017 AVEC challenge, only the training and development sets of the DAIC-WOZ database are available. However, performing both optimization and testing on the development set will lead to significant overfitting on the development set. Therefore, we adopt the following test strategies for our experiments:

\begin{enumerate}

\item{\bf 10-fold stratified cross-validation (CV)}: In this test strategy, the training set and development set are concatenated together and then divided into 10 folds in a stratified manner. Each time, one fold is used for testing and another 9 folds are used for training. Note that the random oversampling and model-dependent feature selection are conducted after the data splitting and only on the training data. Since it is not meaningful to conduct CFS feature selection using cross-validation, the model-independent feature selection is conducted on the entire training and development set, which will lead to an over-optimistic estimate on the test result, but will not affect other hyper-parameter selections.  Thus, we believe this is the fairest way of testing. All optimizations, including model selection, hyper-parameter tuning, and feature selection, are performed according to the results of CV. 

\item {\bf Test on the development set (Dev)}: In this strategy, we train the model using the official training set and test on the official development set. In order to avoid reporting over-optimistic results on the development set, we do not conduct any optimization for the development set. Instead, we find the best model, hyper-parameters, and feature numbers in the CV test and use them to build the model on the training set. 

\item {\bf Test on the testing set (Test)}: In this strategy, we train the model using the official training and development set and test on the official test set. This is because we want to use all available data for training to increase the model robustness. Again, all parameters used in building the model are selected in the CV test.

\end{enumerate}

\subsection{Metrics}
In this work, we report four metrics for each test strategy mentioned above: 1) \textbf{Root mean square error (RMSE)} is the challenge target; therefore, all optimizations, including model selection and feature selection, are performed according to this metric. 2) \textbf{Mean absolute error (MAE)} is another metric reported by the official baseline~\cite{avec2017baseline} and we use it together with RMSE to analyze the difference between ground truth and prediction. 3) \textbf{Pearson correlation coefficient (CC)} is an important metric to evaluate the regression performance, which can reflect the linear correlation between ground truth and prediction. 4) \textbf{F1-score}  measures the performance of binary depression classification, i.e., a subject is depressed when the PHQ-8 score is greater than or equal to 10 and non-depressed otherwise.

\begin{table*}[htp]
\centering
\caption{Result of the depression regression experiment  \protect\footnotemark[1]}
\label{tab:result}
\begin{tabular}{@{}lcccccccccccc@{}}
\toprule
                   &               & RMSE          &      &               & MAE           &      &               & CC            &      &               & F1-Score      &      \\ \midrule
                   & CV            & Dev           & Test & CV            & Dev           & Test & CV            & Dev           & Test & CV            & Dev           & Test \\
Basic baseline     & 5.84          & 6.57          & /    & 4.81          & 5.50          & /    & -0.35        & 0.00          & /    & 0.00          & 0.00          & /    \\
Challenge baseline & /             & 7.13          & 6.97 & /             & 5.88          & 6.12 & /             & /             & /    & /             & /             & /    \\
Context-unaware baseline       & 5.55          & 5.02          & /    & 4.56          & 4.42          & /    & 0.45          & 0.69          & /    & 0.58          & 0.67          & /    \\
Proposed method   & \textbf{3.68} & \textbf{3.54} & \textbf{4.99} & \textbf{2.94} & \textbf{2.77} & \textbf{3.96} & \textbf{0.78} & \textbf{0.87} & /    & \textbf{0.80} & \textbf{0.70} & \textbf{0.60} \\ \bottomrule
\end{tabular}
\end{table*}

\subsection{Baseline}

We compare the proposed method with the following baseline methods:

\begin{enumerate}

\item{\bf Basic Baseline}, where the model constantly predicts the mean PHQ-8 score of the training set. This is a very basic baseline that any workable regression algorithm should outperform.

\item {\bf Challenge Baseline}, which is the official baseline provided in~\cite{avec2017baseline}. This baseline uses a random forest regressor (number of trees = 10) on the audio and video features extracted by the COVAREP toolkit~\cite{degottex2014covarep} and OpenFace toolkit~\cite{baltruvsaitis2016openface}. Regression is performed on a frame-wise basis and the temporal fusion over the interview is compressed by averaging the outputs over the entire interview. Fusion of audio and video modalities is performed by averaging the regression outputs of the unimodal result. In~\cite{avec2017baseline}, the authors present the results of audio unimodal, video unimodal, and audio/video multimodal solutions using this baseline approach, where video unimodal has the best performance. Therefore, we use the results of the video unimodal solution for comparison.

\item {\bf Context-unaware Baseline}
Since the proposed method and the official challenge baseline method have a lot of differences in terms of features, regression model, and class balancing, it is hard to judge which factor cause any performance gap. Therefore, we use this baseline to check the effectiveness of context-aware analysis. This baseline method is exactly the same as the proposed method (i.e., the same audio, video, and LIWC features are extracted, the same feature selection algorithms are used, and the regression model is selected from the same grid), except topic modeling is not used. The differences are that features are extracted and averaged over the entire interview (instead of a topic-wise manner) and that topic related features (topic presence and key topic features) are not included.

\item {\bf Proposed Method}, as described in Section~\ref{sec:proposed}.

\end{enumerate}

\section{Evaluation and Discussion}

\subsection{Overall Performance}
Through a grid search in the CV test, we selected the best regression model (SGD regressor) and the best number of features (46). We then use these settings on the Dev and Test experiments. The results of these experiments are shown is Table~\ref{tab:result}. We observe that the proposed method achieves the best performance for all metrics and test strategies. Further, we find that the proposed method performs significantly better than the context-unaware baseline, which demonstrates the effectiveness of context-aware analysis. In addition, we observe that the performance of the proposed method on the test set is worse than that on the development set and cross validation. This is because the model-independent CFS feature selection is conducted not in a cross-validation manner, but instead on the entire training and development set, since it is meaningless to conduct CFS in a cross-validation manner. However, the performance of the proposed method is still much higher than the challenge baseline on the test data.

\subsection{Selected Feature Analysis}

\begin{figure}[htp]
  \centering
  \includegraphics[width=8.4cm]{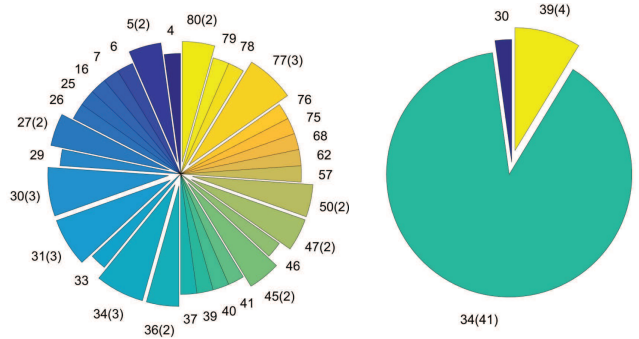}
  \caption{Distribution of topics corresponding to the selected features (count in parentheses). Left: proposed feature selection algorithm. Right: baseline feature selection algorithm.}
  \label{fig:topic}
\end{figure}

It is very interesting to see which features are actually selected and useful in depression detection. In our feature building scheme, each feature corresponds to one topic and one feature category. As shown in Figure~\ref{fig:topic} (left), from the perspective of the topics involved, we observe that 31 topics out of the total 83 topics are involved, in which the most frequent topics corresponding to the selected features are topic 30: how\_long\_diagnose, 31: guilty, 34: seek\_help, and 77: depression. Further, we observe that our approach uses a variety of topics that seem not closely related to depression from humans' perspective such as topic 6: advice\_ago and 16: enjoy\_travel. In addition, to check the effectiveness of the proposed two-step feature selection algorithm, we compare it with a baseline feature selection algorithm that only consists of step 2 of the proposed method, which only considers the score of each feature individually. As shown in Figure~\ref{fig:topic} (right), the feature vector selected by the baseline feature selection algorithm only includes three topics: 30: how\_long\_diagnose, 34: seek\_help, and 39: still\_therapy. We conduct an experiment using this feature vector and find that the result (RMSE: 5.60) is much worse than the result of the proposed approach (RMSE: 4.99) on the test set. This demonstrates 
that the proposed two-step feature selection algorithm is able to discover independent features and to improve the result. While it is possible that topics 30, 34, and 39 are the closest related topics of depression, taking more topics into consideration can lead to a more precise prediction. We believe that it is also an advantage of the proposed method over a clinician's analysis, because for a clinician it is very hard to observe and model such a large volume of factors in the interview.

\begin{figure}[htp]
  \centering
  \includegraphics[width=4.6cm]{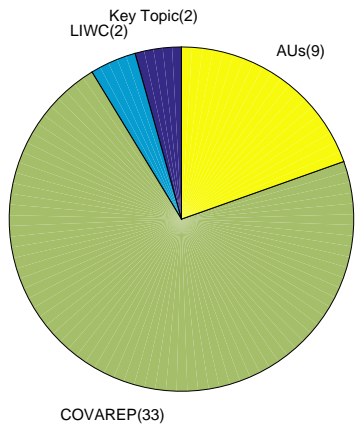}
  \caption{Distribution of feature categories corresponding to the selected features (count in parentheses).}
  \label{fig:cate}
\end{figure}

From the perspective of feature categories involved, we observe that the selected feature set involves LIWC features, key topic semantic features, COVAREP audio features, and AUs video features. The two key topics involved are 78: easy\_sleep and 80: feeling\_lately. The gender feature, topic presence feature, and the formant features are not involved. A complete pie chart of the distribution of feature categories corresponding to the selected feature set is shown in Figure~\ref{fig:cate}. 

\footnotetext[1]{Due to the limited number of test attempts allowed in the 2017 AVEC, we are not able to provide the results on the test set for the baseline approaches. The challenge baseline paper~\cite{avec2017baseline} does not include test results of CC and F1-score and does also not include results tested in CV manner.}

\subsection{Regression Model and Feature Number Analysis}

\begin{figure}[htp]
  \centering
  \includegraphics[width=7.5cm]{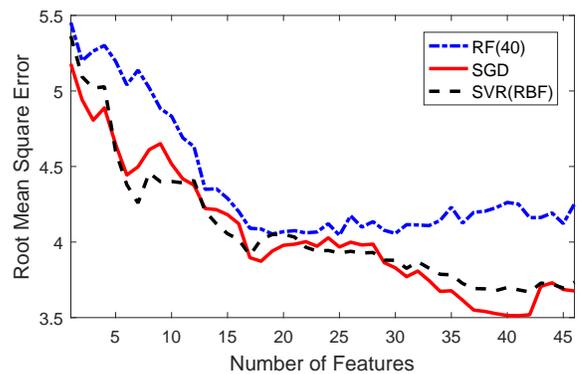}
  \caption{The relationship between RMSE and number of features and regression models.}
  \label{fig:RMSE}
\end{figure}

\begin{figure}[htp]
  \centering
  \includegraphics[width=7.5cm]{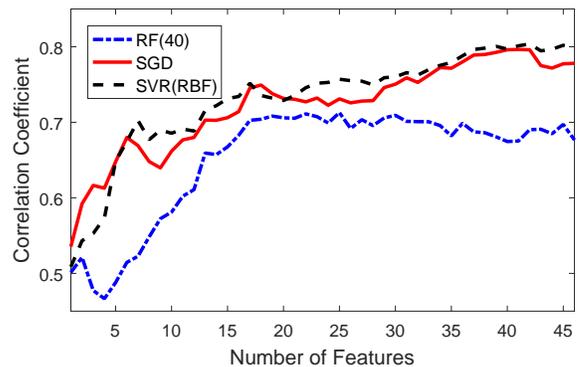}
  \caption{The relationship between CC and number of features and regression models.}
  \label{fig:CC}
\end{figure}

Two important hyperparameters in the proposed method are the number of features and the regression model. Thus, we performed a grid search in cross validation manner for the following regression models: random forest regression (RF) (number of trees: 1, 10, 20, 30, 40, 50, 100, 200), SGD regression, and support vector regression (SVR) (kernel: linear, polynomial, and RBF), and the following feature numbers: 1-46 (the total number of features in the subset selected by the first round CFS feature selection is 46). The relationship between regression performance and hyperparameters is shown in Figures~\ref{fig:RMSE} and~\ref{fig:CC}. For clarity, we only plot the top 3 regression models with the best performance. 

We observe that when the feature number is small, the random forest regressor (tree number = 40), SGD regressor, and SVR (RBF) regressor perform similarly. However, with the increase of feature numbers, SGD and SVR models continually improve their performance while the random forest model stops improving much earlier. The SGD and SVR regressor have close performance, while the SGD regressor has a little bit lower RMSE than SVR.  Though the lowest RMSE is achieved when the feature number is 41, we believe it is more likely to be a fluctuation in the CV test and therefore we choose the feature number of 46, because we prefer to use more features to build a more discriminative model. The experiment shows that using 46 features (RMSE: 4.99) yields a better performance than using 41 features (RMSE:5.22) on the test set.

\section{Conclusions}

Major depressive disorder is a widespread mental disorder and accurate detection will be essential for targeted intervention and treatment. In this challenge, participants are asked to build a model predicting the depression levels based on the audio, video, and text of an interview ranging between 7-33 minutes. Since averaging features over the entire interview will lose most temporal details, how to discover, capture, and preserve important temporal details for such long interviews are significant challenges. Therefore, we propose a novel topic modeling based approach to perform context-aware analysis. Our experiments show that the proposed approach performs significantly better than context-unaware method and the challenge baseline for all metrics. In addition, by analyzing the features selected by the machine learning algorithm, we found that our approach has the ability to discover a variety of temporal features that have underlying relationship with depression and further to build model on them, which is a task that is difficult to perform by humans.

\bibliographystyle{ACM-Reference-Format}
\bibliography{sigchi-a.bib} 


\begin{thebibliography}{00}


\ifx \showCODEN    \undefined \def \showCODEN     #1{\unskip}     \fi
\ifx \showDOI      \undefined \def \showDOI       #1{#1}\fi
\ifx \showISBNx    \undefined \def \showISBNx     #1{\unskip}     \fi
\ifx \showISBNxiii \undefined \def \showISBNxiii  #1{\unskip}     \fi
\ifx \showISSN     \undefined \def \showISSN      #1{\unskip}     \fi
\ifx \showLCCN     \undefined \def \showLCCN      #1{\unskip}     \fi
\ifx \shownote     \undefined \def \shownote      #1{#1}          \fi
\ifx \showarticletitle \undefined \def \showarticletitle #1{#1}   \fi
\ifx \showURL      \undefined \def \showURL       {\relax}        \fi
\providecommand\bibfield[2]{#2}
\providecommand\bibinfo[2]{#2}
\providecommand\natexlab[1]{#1}
\providecommand\showeprint[2][]{arXiv:#2}

\bibitem[\protect\citeauthoryear{Baltru{\v{s}}aitis, Robinson, and
  Morency}{Baltru{\v{s}}aitis et~al\mbox{.}}{2016}]%
        {baltruvsaitis2016openface}
\bibfield{author}{\bibinfo{person}{Tadas Baltru{\v{s}}aitis},
  \bibinfo{person}{Peter Robinson}, {and} \bibinfo{person}{Louis-Philippe
  Morency}.} \bibinfo{year}{2016}\natexlab{}.
\newblock \showarticletitle{Openface: an open source facial behavior analysis
  toolkit}. In \bibinfo{booktitle}{{\em Applications of Computer Vision (WACV),
  2016 IEEE Winter Conference on}}. IEEE, \bibinfo{pages}{1--10}.
\newblock


\bibitem[\protect\citeauthoryear{Blei, Ng, and Jordan}{Blei
  et~al\mbox{.}}{2003}]%
        {blei2003latent}
\bibfield{author}{\bibinfo{person}{David~M Blei}, \bibinfo{person}{Andrew~Y
  Ng}, {and} \bibinfo{person}{Michael~I Jordan}.}
  \bibinfo{year}{2003}\natexlab{}.
\newblock \showarticletitle{Latent dirichlet allocation}.
\newblock \bibinfo{journal}{{\em Journal of machine Learning research\/}}
  \bibinfo{volume}{3}, \bibinfo{number}{Jan} (\bibinfo{year}{2003}),
  \bibinfo{pages}{993--1022}.
\newblock


\bibitem[\protect\citeauthoryear{Busso, Bulut, Lee, Kazemzadeh, Mower, Kim,
  Chang, Lee, and Narayanan}{Busso et~al\mbox{.}}{2008}]%
        {busso2008iemocap}
\bibfield{author}{\bibinfo{person}{Carlos Busso}, \bibinfo{person}{Murtaza
  Bulut}, \bibinfo{person}{Chi-Chun Lee}, \bibinfo{person}{Abe Kazemzadeh},
  \bibinfo{person}{Emily Mower}, \bibinfo{person}{Samuel Kim},
  \bibinfo{person}{Jeannette~N Chang}, \bibinfo{person}{Sungbok Lee}, {and}
  \bibinfo{person}{Shrikanth~S Narayanan}.} \bibinfo{year}{2008}\natexlab{}.
\newblock \showarticletitle{IEMOCAP: Interactive emotional dyadic motion
  capture database}.
\newblock \bibinfo{journal}{{\em Language resources and evaluation\/}}
  \bibinfo{volume}{42}, \bibinfo{number}{4} (\bibinfo{year}{2008}),
  \bibinfo{pages}{335}.
\newblock


\bibitem[\protect\citeauthoryear{Degottex, Kane, Drugman, Raitio, and
  Scherer}{Degottex et~al\mbox{.}}{2014}]%
        {degottex2014covarep}
\bibfield{author}{\bibinfo{person}{Gilles Degottex}, \bibinfo{person}{John
  Kane}, \bibinfo{person}{Thomas Drugman}, \bibinfo{person}{Tuomo Raitio},
  {and} \bibinfo{person}{Stefan Scherer}.} \bibinfo{year}{2014}\natexlab{}.
\newblock \showarticletitle{COVAREP: A collaborative voice analysis repository
  for speech technologies}. In \bibinfo{booktitle}{{\em Acoustics, Speech and
  Signal Processing (ICASSP), 2014 IEEE International Conference on}}. IEEE,
  \bibinfo{pages}{960--964}.
\newblock


\bibitem[\protect\citeauthoryear{DeVault, Artstein, Benn, Dey, Fast, Gainer,
  Georgila, Gratch, Hartholt, Lhommet, et~al\mbox{.}}{DeVault
  et~al\mbox{.}}{2014}]%
        {devault2014simsensei}
\bibfield{author}{\bibinfo{person}{David DeVault}, \bibinfo{person}{Ron
  Artstein}, \bibinfo{person}{Grace Benn}, \bibinfo{person}{Teresa Dey},
  \bibinfo{person}{Ed Fast}, \bibinfo{person}{Alesia Gainer},
  \bibinfo{person}{Kallirroi Georgila}, \bibinfo{person}{Jon Gratch},
  \bibinfo{person}{Arno Hartholt}, \bibinfo{person}{Margaux Lhommet},
  {et~al\mbox{.}}} \bibinfo{year}{2014}\natexlab{}.
\newblock \showarticletitle{SimSensei Kiosk: A virtual human interviewer for
  healthcare decision support}. In \bibinfo{booktitle}{{\em Proceedings of the
  2014 international conference on Autonomous agents and multi-agent systems}}.
  International Foundation for Autonomous Agents and Multiagent Systems,
  \bibinfo{pages}{1061--1068}.
\newblock


\bibitem[\protect\citeauthoryear{Fava and Kendler}{Fava and Kendler}{2000}]%
        {fava2000major}
\bibfield{author}{\bibinfo{person}{Maurizio Fava} {and}
  \bibinfo{person}{Kenneth~S Kendler}.} \bibinfo{year}{2000}\natexlab{}.
\newblock \showarticletitle{Major depressive disorder}.
\newblock \bibinfo{journal}{{\em Neuron\/}} \bibinfo{volume}{28},
  \bibinfo{number}{2} (\bibinfo{year}{2000}), \bibinfo{pages}{335--341}.
\newblock


\bibitem[\protect\citeauthoryear{Gratch, Artstein, Lucas, Stratou, Scherer,
  Nazarian, Wood, Boberg, DeVault, Marsella, et~al\mbox{.}}{Gratch
  et~al\mbox{.}}{2014}]%
        {gratch2014distress}
\bibfield{author}{\bibinfo{person}{Jonathan Gratch}, \bibinfo{person}{Ron
  Artstein}, \bibinfo{person}{Gale~M Lucas}, \bibinfo{person}{Giota Stratou},
  \bibinfo{person}{Stefan Scherer}, \bibinfo{person}{Angela Nazarian},
  \bibinfo{person}{Rachel Wood}, \bibinfo{person}{Jill Boberg},
  \bibinfo{person}{David DeVault}, \bibinfo{person}{Stacy Marsella},
  {et~al\mbox{.}}} \bibinfo{year}{2014}\natexlab{}.
\newblock \showarticletitle{The Distress Analysis Interview Corpus of human and
  computer interviews}. In \bibinfo{booktitle}{{\em LREC}}.
  \bibinfo{pages}{3123--3128}.
\newblock


\bibitem[\protect\citeauthoryear{Hall}{Hall}{1998}]%
        {hall1998correlation}
\bibfield{author}{\bibinfo{person}{MA Hall}.} \bibinfo{year}{1998}\natexlab{}.
\newblock \showarticletitle{Correlation-based feature subset selection for
  machine learning}.
\newblock \bibinfo{journal}{{\em Thesis submitted in partial fulfillment of the
  requirements of the degree of Doctor of Philosophy at the University of
  Waikato\/}} (\bibinfo{year}{1998}).
\newblock


\bibitem[\protect\citeauthoryear{Hong and Davison}{Hong and Davison}{2010}]%
        {hong2010empirical}
\bibfield{author}{\bibinfo{person}{Liangjie Hong} {and}
  \bibinfo{person}{Brian~D Davison}.} \bibinfo{year}{2010}\natexlab{}.
\newblock \showarticletitle{Empirical study of topic modeling in twitter}. In
  \bibinfo{booktitle}{{\em Proceedings of the first workshop on social media
  analytics}}. ACM, \bibinfo{pages}{80--88}.
\newblock


\bibitem[\protect\citeauthoryear{Kroenke, Strine, Spitzer, Williams, Berry, and
  Mokdad}{Kroenke et~al\mbox{.}}{2009}]%
        {kroenke2009phq}
\bibfield{author}{\bibinfo{person}{Kurt Kroenke}, \bibinfo{person}{Tara~W
  Strine}, \bibinfo{person}{Robert~L Spitzer}, \bibinfo{person}{Janet~BW
  Williams}, \bibinfo{person}{Joyce~T Berry}, {and} \bibinfo{person}{Ali~H
  Mokdad}.} \bibinfo{year}{2009}\natexlab{}.
\newblock \showarticletitle{The PHQ-8 as a measure of current depression in the
  general population}.
\newblock \bibinfo{journal}{{\em Journal of affective disorders\/}}
  \bibinfo{volume}{114}, \bibinfo{number}{1} (\bibinfo{year}{2009}),
  \bibinfo{pages}{163--173}.
\newblock


\bibitem[\protect\citeauthoryear{Liu, Wu, and Zhou}{Liu et~al\mbox{.}}{2009}]%
        {liu2009exploratory}
\bibfield{author}{\bibinfo{person}{Xu-Ying Liu}, \bibinfo{person}{Jianxin Wu},
  {and} \bibinfo{person}{Zhi-Hua Zhou}.} \bibinfo{year}{2009}\natexlab{}.
\newblock \showarticletitle{Exploratory undersampling for class-imbalance
  learning}.
\newblock \bibinfo{journal}{{\em IEEE Transactions on Systems, Man, and
  Cybernetics, Part B (Cybernetics)\/}} \bibinfo{volume}{39},
  \bibinfo{number}{2} (\bibinfo{year}{2009}), \bibinfo{pages}{539--550}.
\newblock


\bibitem[\protect\citeauthoryear{Mei, Cai, Zhang, and Zhai}{Mei
  et~al\mbox{.}}{2008}]%
        {mei2008topic}
\bibfield{author}{\bibinfo{person}{Qiaozhu Mei}, \bibinfo{person}{Deng Cai},
  \bibinfo{person}{Duo Zhang}, {and} \bibinfo{person}{ChengXiang Zhai}.}
  \bibinfo{year}{2008}\natexlab{}.
\newblock \showarticletitle{Topic modeling with network regularization}. In
  \bibinfo{booktitle}{{\em Proceedings of the 17th international conference on
  World Wide Web}}. ACM, \bibinfo{pages}{101--110}.
\newblock


\bibitem[\protect\citeauthoryear{Nasir, Jati, Shivakumar,
  Nallan~Chakravarthula, and Georgiou}{Nasir et~al\mbox{.}}{2016}]%
        {nasir2016multimodal}
\bibfield{author}{\bibinfo{person}{Md Nasir}, \bibinfo{person}{Arindam Jati},
  \bibinfo{person}{Prashanth~Gurunath Shivakumar}, \bibinfo{person}{Sandeep
  Nallan~Chakravarthula}, {and} \bibinfo{person}{Panayiotis Georgiou}.}
  \bibinfo{year}{2016}\natexlab{}.
\newblock \showarticletitle{Multimodal and multiresolution depression detection
  from speech and facial landmark features}. In \bibinfo{booktitle}{{\em
  Proceedings of the 6th International Workshop on Audio/Visual Emotion
  Challenge}}. ACM, \bibinfo{pages}{43--50}.
\newblock


\bibitem[\protect\citeauthoryear{Pampouchidou, Simantiraki, Fazlollahi,
  Pediaditis, Manousos, Roniotis, Giannakakis, Meriaudeau, Simos, Marias,
  et~al\mbox{.}}{Pampouchidou et~al\mbox{.}}{2016}]%
        {pampouchidou2016depression}
\bibfield{author}{\bibinfo{person}{Anastasia Pampouchidou},
  \bibinfo{person}{Olympia Simantiraki}, \bibinfo{person}{Amir Fazlollahi},
  \bibinfo{person}{Matthew Pediaditis}, \bibinfo{person}{Dimitris Manousos},
  \bibinfo{person}{Alexandros Roniotis}, \bibinfo{person}{Georgios
  Giannakakis}, \bibinfo{person}{Fabrice Meriaudeau},
  \bibinfo{person}{Panagiotis Simos}, \bibinfo{person}{Kostas Marias},
  {et~al\mbox{.}}} \bibinfo{year}{2016}\natexlab{}.
\newblock \showarticletitle{Depression Assessment by Fusing High and Low Level
  Features from Audio, Video, and Text}. In \bibinfo{booktitle}{{\em
  Proceedings of the 6th International Workshop on Audio/Visual Emotion
  Challenge}}. ACM, \bibinfo{pages}{27--34}.
\newblock


\bibitem[\protect\citeauthoryear{Pennebaker, Boyd, Jordan, and
  Blackburn}{Pennebaker et~al\mbox{.}}{2015}]%
        {pennebaker2015development}
\bibfield{author}{\bibinfo{person}{James~W Pennebaker}, \bibinfo{person}{Ryan~L
  Boyd}, \bibinfo{person}{Kayla Jordan}, {and} \bibinfo{person}{Kate
  Blackburn}.} \bibinfo{year}{2015}\natexlab{}.
\newblock \bibinfo{booktitle}{{\em The development and psychometric properties
  of LIWC2015}}.
\newblock \bibinfo{type}{{T}echnical {R}eport}.
\newblock


\bibitem[\protect\citeauthoryear{Resnik, Garron, and Resnik}{Resnik
  et~al\mbox{.}}{2013}]%
        {resnik2013using}
\bibfield{author}{\bibinfo{person}{Philip Resnik}, \bibinfo{person}{Anderson
  Garron}, {and} \bibinfo{person}{Rebecca Resnik}.}
  \bibinfo{year}{2013}\natexlab{}.
\newblock \showarticletitle{Using topic modeling to improve prediction of
  neuroticism and depression}. In \bibinfo{booktitle}{{\em Proceedings of the
  2013 Conference on Empirical Methods in Natural}}. Association for
  Computational Linguistics, \bibinfo{pages}{1348--1353}.
\newblock


\bibitem[\protect\citeauthoryear{Ringeval, Schuller, Valstar, Gratch, Cowie,
  Scherer, Mozgai, Cummins, Schmitt, and Pantic}{Ringeval
  et~al\mbox{.}}{2017}]%
        {avec2017baseline}
\bibfield{author}{\bibinfo{person}{Fabien Ringeval}, \bibinfo{person}{Bj{\"o}rn
  Schuller}, \bibinfo{person}{Michel Valstar}, \bibinfo{person}{Jonathan
  Gratch}, \bibinfo{person}{Roddy Cowie}, \bibinfo{person}{Stefan Scherer},
  \bibinfo{person}{Sharon Mozgai}, \bibinfo{person}{Nicholas Cummins},
  \bibinfo{person}{Maximilian Schmitt}, {and} \bibinfo{person}{Maja Pantic}.}
  \bibinfo{year}{2017}\natexlab{}.
\newblock \showarticletitle{AVEC 2017: Real-life Depression, and Affect
  Recognition Workshop and Challenge}. In \bibinfo{booktitle}{{\em Proceedings
  of the 7th International Workshop on Audio/Visual Emotion Challenge}}. ACM,
  \bibinfo{pages}{1--8}.
\newblock


\bibitem[\protect\citeauthoryear{Spijker, De~Graaf, Bijl, Beekman, Ormel, and
  Nolen}{Spijker et~al\mbox{.}}{2002}]%
        {spijker2002duration}
\bibfield{author}{\bibinfo{person}{JAN Spijker}, \bibinfo{person}{Ron
  De~Graaf}, \bibinfo{person}{Rob~V Bijl}, \bibinfo{person}{Aartjan~TF
  Beekman}, \bibinfo{person}{Johan Ormel}, {and} \bibinfo{person}{Willem~A
  Nolen}.} \bibinfo{year}{2002}\natexlab{}.
\newblock \showarticletitle{Duration of major depressive episodes in the
  general population: results from The Netherlands Mental Health Survey and
  Incidence Study (NEMESIS)}.
\newblock \bibinfo{journal}{{\em The British journal of psychiatry\/}}
  \bibinfo{volume}{181}, \bibinfo{number}{3} (\bibinfo{year}{2002}),
  \bibinfo{pages}{208--213}.
\newblock


\bibitem[\protect\citeauthoryear{Valstar, Gratch, Schuller, Ringeval, Lalanne,
  Torres~Torres, Scherer, Stratou, Cowie, and Pantic}{Valstar
  et~al\mbox{.}}{2016}]%
        {valstar2016avec}
\bibfield{author}{\bibinfo{person}{Michel Valstar}, \bibinfo{person}{Jonathan
  Gratch}, \bibinfo{person}{Bj{\"o}rn Schuller}, \bibinfo{person}{Fabien
  Ringeval}, \bibinfo{person}{Dennis Lalanne}, \bibinfo{person}{Mercedes
  Torres~Torres}, \bibinfo{person}{Stefan Scherer}, \bibinfo{person}{Giota
  Stratou}, \bibinfo{person}{Roddy Cowie}, {and} \bibinfo{person}{Maja
  Pantic}.} \bibinfo{year}{2016}\natexlab{}.
\newblock \showarticletitle{Avec 2016: Depression, mood, and emotion
  recognition workshop and challenge}. In \bibinfo{booktitle}{{\em Proceedings
  of the 6th International Workshop on Audio/Visual Emotion Challenge}}. ACM,
  \bibinfo{pages}{3--10}.
\newblock


\bibitem[\protect\citeauthoryear{Wang and Blei}{Wang and Blei}{2011}]%
        {wang2011collaborative}
\bibfield{author}{\bibinfo{person}{Chong Wang} {and} \bibinfo{person}{David~M
  Blei}.} \bibinfo{year}{2011}\natexlab{}.
\newblock \showarticletitle{Collaborative topic modeling for recommending
  scientific articles}. In \bibinfo{booktitle}{{\em Proceedings of the 17th ACM
  SIGKDD international conference on Knowledge discovery and data mining}}.
  ACM, \bibinfo{pages}{448--456}.
\newblock


\bibitem[\protect\citeauthoryear{Williamson, Godoy, Cha, Schwarzentruber,
  Khorrami, Gwon, Kung, Dagli, and Quatieri}{Williamson et~al\mbox{.}}{2016}]%
        {williamson2016detecting}
\bibfield{author}{\bibinfo{person}{James~R Williamson},
  \bibinfo{person}{Elizabeth Godoy}, \bibinfo{person}{Miriam Cha},
  \bibinfo{person}{Adrianne Schwarzentruber}, \bibinfo{person}{Pooya Khorrami},
  \bibinfo{person}{Youngjune Gwon}, \bibinfo{person}{Hsiang-Tsung Kung},
  \bibinfo{person}{Charlie Dagli}, {and} \bibinfo{person}{Thomas~F Quatieri}.}
  \bibinfo{year}{2016}\natexlab{}.
\newblock \showarticletitle{Detecting Depression using Vocal, Facial and
  Semantic Communication Cues}. In \bibinfo{booktitle}{{\em Proceedings of the
  6th International Workshop on Audio/Visual Emotion Challenge}}. ACM,
  \bibinfo{pages}{11--18}.
\newblock


\bibitem[\protect\citeauthoryear{Yang, Jiang, He, Pei, Oveneke, and Sahli}{Yang
  et~al\mbox{.}}{2016}]%
        {yang2016decision}
\bibfield{author}{\bibinfo{person}{Le Yang}, \bibinfo{person}{Dongmei Jiang},
  \bibinfo{person}{Lang He}, \bibinfo{person}{Ercheng Pei},
  \bibinfo{person}{Meshia~C{\'e}dric Oveneke}, {and} \bibinfo{person}{Hichem
  Sahli}.} \bibinfo{year}{2016}\natexlab{}.
\newblock \showarticletitle{Decision Tree Based Depression Classification from
  Audio Video and Language Information}. In \bibinfo{booktitle}{{\em
  Proceedings of the 6th International Workshop on Audio/Visual Emotion
  Challenge}}. ACM, \bibinfo{pages}{89--96}.
\newblock


\end{thebibliography}

\end{document}